\documentclass[10pt,twocolumn,letterpaper]{article}

\usepackage[pagenumbers]{cvpr}
\usepackage{placeins}

\usepackage{booktabs}
\usepackage{multirow}
\usepackage{graphicx}

\newcommand{\qheading}[1]{\noindent\mbox{\textbf{#1}\;}}

\definecolor{cvprblue}{rgb}{0.21,0.49,0.74}
\usepackage[pagebackref,breaklinks,colorlinks,allcolors=cvprblue]{hyperref}

\def\confName{CVPR}
\def\confYear{2027}

\title{Ego2Real: Learning Bimanual Dexterous Manipulation from Egocentric Visuo-Tactile Demonstrations}

\author{
\begin{tabular}[t]{@{}c@{}}
{\large Ange Bao$^{*1,3}$\quad
Haishan Zhang$^{*2,3}$\quad
Lingzi Xie$^{3}$\quad
Shuang Li$^{3}$}\\
{\large Pei Zhao$^{1}$\quad
Ling Shi$^{2}$\quad
Bidan Huang$^{\dagger 3}$}\\[4pt]
{\normalsize $^1$Zhejiang University
\qquad
$^2$The Hong Kong University of Science and Technology
\qquad
$^3$AGILINK}
\end{tabular}
}

\begin{document}

\setlength{\floatsep}{4pt}
\setlength{\textfloatsep}{10pt plus 2pt minus 2pt}

\twocolumn[{%
\renewcommand\twocolumn[1][]{#1}%
\maketitle

\begin{center}
  \includegraphics[width=\textwidth]{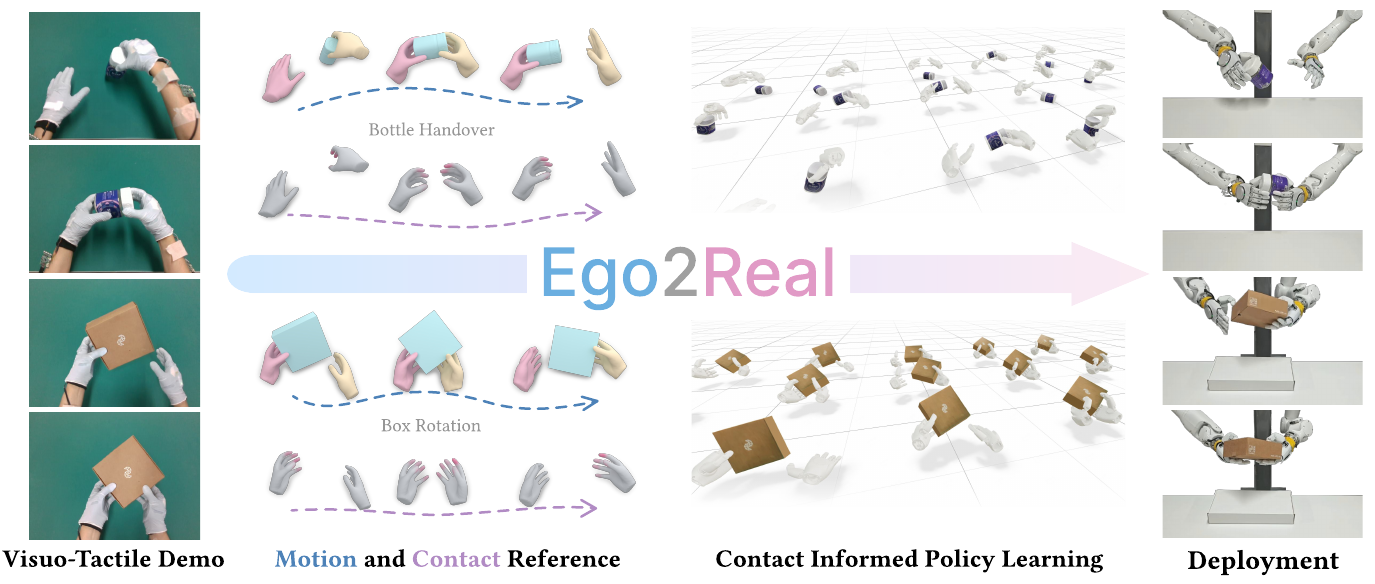}
  \captionof{figure}{\textbf{Overview of Ego2Real.}
  From egocentric demonstrations recorded with an RGB-D camera and tactile gloves,
  Ego2Real reconstructs bimanual hand--object interactions with tactile-guided refinement,
  yielding both a motion reference and a finger-wise contact reference.
  These references supervise contact-informed residual RL for robotic hands in simulation,
  and the resulting policies are deployed on real bimanual robots.}
  \label{fig:teaser}
\end{center}%
}]

\begingroup
\renewcommand{\thefootnote}{\fnsymbol{footnote}}
\footnotetext[1]{Equal contribution; work done while interning at AGILINK.}
\footnotetext[2]{AGILINK team lead.}
\endgroup

\begin{abstract}
Learning dexterous manipulation from egocentric human demonstrations offers a scalable path toward robot skill acquisition. 
Vision alone, however, provides limited information about physical contacts and the demonstrator's force intent, leading to ambiguities in interpreting hand-object interactions. 
Capturing human touch via tactile sensing is therefore crucial for reliable human-to-robot dexterous skill transfer.
We present \textbf{Ego2Real}, a framework that leverages egocentric demonstrations and tactile gloves to transfer bimanual dexterous manipulation skills to robotic hands.
Ego2Real incorporates tactile supervision at two complementary stages: 
(1) tactile-guided hand-object interaction reconstruction, which uses measured tactile signals to refine hand poses and contact geometry; 
and (2) contact-informed reinforcement learning, which translates human contact patterns into robotic contact rewards. 
By combining refined motion references with physical contact rewards during training, Ego2Real guides anthropomorphic robotic hands to emulate demonstrated contact behaviors while adapting to their own morphology and actuation.
Experiments show that incorporating tactile supervision into both reconstruction and policy learning yields higher manipulation success rates, underscoring the importance of enriching egocentric demonstrations with tactile observations to improve their quality and transferability.

\end{abstract}

\section{Introduction}
\label{sec:intro}

Dexterous manipulation requires precise finger coordination and controlled hand--object interactions.
Learning from human demonstrations offers a natural way for robots to acquire such skills.
Egocentric recordings are particularly appealing~\cite{egomatrix,unidex,egoscale} because they allow everyday manipulation to be captured with lightweight setups and allow demonstrators to perform tasks naturally with their hands.
Such activities usually involve two hands working cooperatively, making bimanual coordination an essential capability to learn from these demonstrations.

Contact plays a central role in this coordination: when two hands interact through a shared object, successful manipulation depends not only on how the hands move, but also on where, when, and how strongly each hand makes and breaks contact.
Yet egocentric visual demonstrations lack direct measurements of these physical interactions.
Prior skill transfer methods often infer contact states and locations from reference hand--object geometry~\cite{maniptrans,chord,dexmachina}.
However, obtaining sufficiently accurate references typically requires high-precision motion capture that restricts collection to instrumented settings, and kinematic geometry alone still cannot reveal actual interaction forces.
Reconstructing hand--object motion directly from videos~\cite{doasido,galatea,chord} removes the need for motion capture, but occlusion and depth ambiguity frequently lead to contact gaps, interpenetration, and unreliable contact estimates.
While tactile gloves~\cite{osmo} have recently enabled direct measurement of human touch for robot learning, most existing approaches~\cite{tactidex,touchbridge,touch2robot} still pair gloves with external motion-capture tracking, leaving their role in egocentric bimanual transfer from noisy visual observations underexplored.
This raises our central question: \textit{how can we combine visual information with directly measured human tactile signals to improve dexterous human-to-robot skill transfer when learning from egocentric data?}

To address this question, we introduce \textbf{Ego2Real} (Fig.~\ref{fig:teaser}), a tactile-guided framework that learns bimanual dexterous manipulation from egocentric demonstrations recorded with tactile gloves and an RGB-D camera.
Given a raw demonstration, Ego2Real extracts a reference encoding a contact-plausible kinematic motion prior, contact states, and force intent, which is used to train residual reinforcement learning policies in simulation for real-world deployment.
Within this pipeline, measured human touch plays a central role at two complementary stages.
First, \textit{tactile-guided hand--object interaction (HOI) reconstruction} maps spatially distributed glove readings onto the hand surface to resolve visual ambiguities, constraining contact geometry and preserving sustained contacts over time.
The refined human trajectories are then retargeted to the robot by preserving hand--object topological interaction relationships, providing a physically plausible motion reference.
Second, \textit{contact-informed reinforcement learning} translates the recorded glove signals into region-wise contact-state, force, and contact-timing rewards that complement motion tracking.
Rather than enforcing rigid trajectory imitation, these rewards guide a residual policy to adapt the retargeted motion to the robot's own morphology and actuation constraints while emulating the demonstrated interaction patterns.
At deployment, the learned policies enable closed-loop execution using tracked object poses and robot proprioception, without requiring robot-side tactile sensing.
Experiments and ablations across simulation and real-world bimanual tasks demonstrate that tactile-guided reconstruction effectively improves downstream skill transfer, while contact-informed policy learning provides additional gains in contact-demanding interactions.

In summary, our contributions are threefold:
\begin{itemize}

    \item \textbf{Bimanual Skill Transfer from Egocentric Visuo-Tactile Demonstrations.}
    We present Ego2Real, a framework that learns dexterous manipulation from synchronized RGB-D and tactile-glove recordings, and show through systematic ablations that incorporating human tactile signals into both reconstruction and policy learning yields complementary gains on real robots.
    \item \textbf{Tactile-Guided HOI Reconstruction.}
    We leverage spatially distributed tactile measurements to constrain hand--object contact geometry and preserve sustained contacts over time, reducing contact ambiguities in visual reconstruction and providing physically plausible references for downstream robot learning.
    \item \textbf{Contact-Informed Reinforcement Learning.}
     We translate human tactile signals into region-wise contact rewards to supervise residual policy learning beyond motion tracking, enabling more human-like contact behaviors and more reliable real-world execution in bimanual tasks.
\end{itemize}

\section{Related Work}

\subsection{Human Demonstrations for Dexterous Manipulation}

Learning dexterous manipulation requires demonstrations that are both scalable to collect and faithful to the underlying physical interaction.
Existing demonstration sources trade off these two properties in different ways.
Real-robot teleoperation~\cite{dexpilot,shaw2024bimanual} yields directly executable demonstrations but requires robot hardware and substantial operator effort, especially for contact-rich behaviors where operators lack high-fidelity tactile feedback.
Simulation-based generation~\cite{dexmimicgen} expands a few source demonstrations but relies on accurate physical models.
Wearable interfaces such as DexUMI~\cite{dexumi} and RealDexUMI~\cite{realdexumi} narrow the human--robot embodiment gap but tie demonstrations to specialized hardware configurations.
Motion-capture datasets~\cite{grab,arctic,oakink2} provide accurate hand--object trajectories and serve as common references for physics-based transfer methods~\cite{maniptrans,dexmachina}, yet their capture systems confine collection to instrumented laboratory environments.
Human videos remove the need for specialized hardware by reconstructing trajectories from everyday, egocentric, or generated videos for simulation-based policy learning~\cite{human2sim2robot,doasido,egomatrix,galatea}, or by retargeting large-scale egocentric data for policy pretraining~\cite{egoscale,unidex}. However, vision-only demonstrations do not directly capture physical contact.

Tactile sensing offers direct evidence of where and when contact occurs.
OSMO~\cite{osmo} introduces an open-source tactile glove compatible with vision-based hand tracking, and transfers human contact signals to a robot equipped with the same glove.
TactiDex~\cite{tactidex} and DexTaG~\cite{dextag} pair tactile gloves with instrumented kinematic capture, using an optical motion-capture system and an electromagnetic motion-capture glove, respectively, to obtain accurate hand--object trajectories, which confines collection to instrumented settings.
In contrast, Ego2Real collects synchronized egocentric RGB-D observations and spatially distributed tactile measurements without external motion capture and recovers hand--object motion from vision.

\subsection{Hand--Object Interaction Reconstruction}
\label{sec:rw_hoi}
Recent methods, including EasyHOI~\cite{liu2025easyhoi}, MagicHOI~\cite{wang2025magichoi}, and AGILE~\cite{shi2026agile}, combine visual observations with foundation-model priors and contact constraints inferred from visual and geometric cues for hand--object reconstruction.
Other approaches explicitly predict contact representations for reconstruction refinement: CPF~\cite{yang2021cpf} uses predicted contact relationships within a contact potential field, while TactoHand~\cite{xi2026tactohand} uses video-predicted contact and proximity fields.
Reconstruction is also a core component of video-based robot learning pipelines: Do as I Do~\cite{doasido}, EgoMatrix~\cite{egomatrix}, and GALATEA~\cite{galatea} reconstruct hand--object interactions from everyday, egocentric, and generated videos, respectively, to obtain references for robot execution.
Across these approaches, contact is inferred from visual or geometric evidence and can remain ambiguous under hand--object occlusion, motivating the use of direct contact measurements.
ContactOpt~\cite{grady2021contactopt} refines hand poses toward predicted contact maps and further shows that thermally measured contact on static grasps can guide the same refinement, while VTacOH~\cite{xu2023vtaco} combines visual point clouds with tactile observations for joint hand--object reconstruction.
ViTaM-D~\cite{yu2025vitamd} uses distributed tactile measurements for hand--object reconstruction with a supervised network and object-category-specific fine-tuning.
Ego2Real uses human tactile signals to guide both bimanual reconstruction and robot policy learning, without task-specific reconstruction training.

\subsection{Human-to-Robot Dexterous Hand Transfer}

Transferring human manipulation skills to dexterous robotic hands is challenging due to differences in hand morphology, kinematics, and contact dynamics. Existing approaches primarily rely on kinematic retargeting, physics-based optimization, and reinforcement learning. Kinematic methods~\cite{dexpilot,toporetarget} align human and robot hand configurations through fingertip correspondence or geometric interaction constraints, while physics-based methods such as SPIDER~\cite{spider} incorporate physical feasibility through sampling-based optimization. Reinforcement-learning-based approaches~\cite{maniptrans,regrind,dexmachina,adadex} further adapt human demonstrations to robot-specific dynamics, enabling physically executable manipulation behaviors.

Contact information is increasingly used to guide cross-embodiment transfer.
CHORD~\cite{chord} guides RL with object-centric contact wrench spaces, while DemoMimic~\cite{demomimic} rewards precise contact around the local geometry of demonstrated contact points.
These contact targets are derived from the reference hand--object geometry and therefore heavily rely on its accuracy.
Human tactile demonstrations provide contact targets independent of this geometry.
DexTaG~\cite{dextag} converts glove tactile readouts into RL rewards, TouchBridge~\cite{touchbridge} incorporates tactile contact into motion retargeting, and TactiDex~\cite{tactidex} uses tactile signals both to post-optimize motion-capture references and to guide RL.
However, whether tactile supervision at the reconstruction and policy-learning stages is complementary when references are recovered from noisy visual observations remains insufficiently explored.
Ego2Real integrates tactile-guided HOI reconstruction with region-wise contact rewards for underactuated robotic hands, and evaluates both stages under matched visual reconstruction noise.

\begin{figure*}
  \centering
  \includegraphics[width=\textwidth]{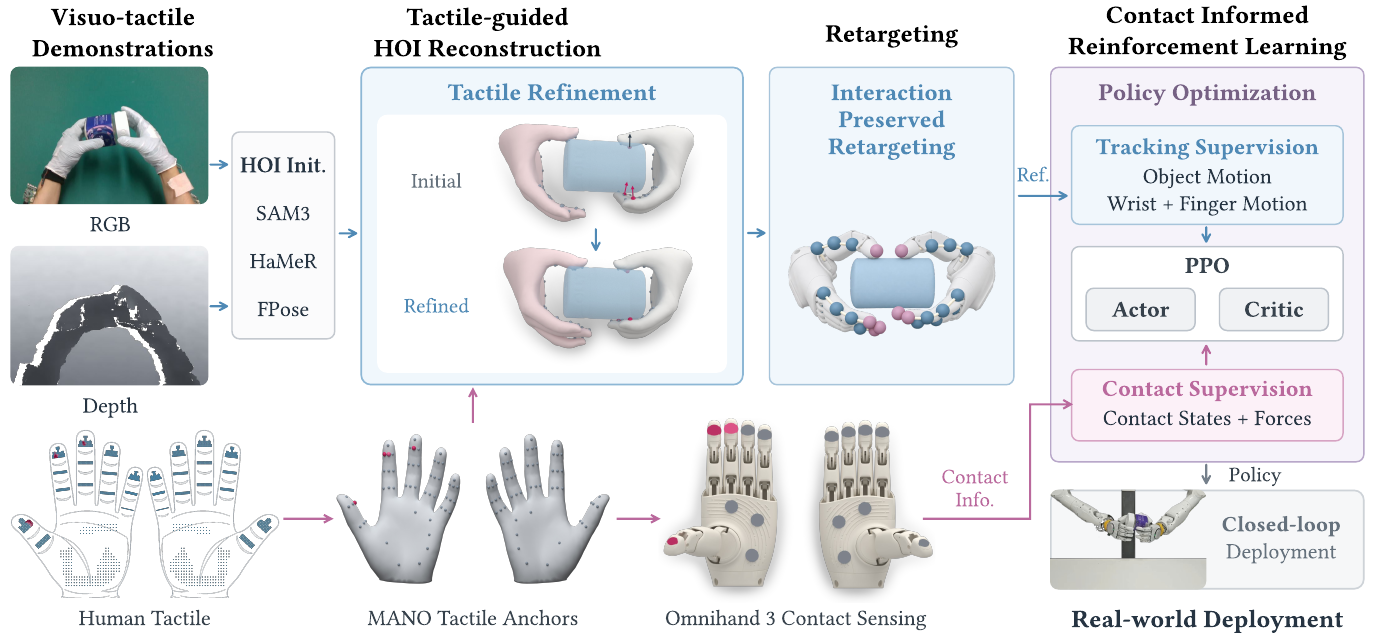}
\caption{\textbf{The Ego2Real Pipeline.}
 RGB-D observations initialize metric hand--object motion, which is refined using human tactile measurements mapped to MANO surface anchors. The refined motion is retargeted to OmniHand 3 while preserving interaction geometry. A policy is then trained with PPO using motion-tracking rewards and contact rewards derived from human contact states and forces. Human tactile measurements thus supervise both HOI reconstruction and policy learning, enabling closed-loop real-world deployment without tactile sensing.}
  \label{fig:ego2real}
\end{figure*}

\section{Method}
\label{sec:method}

Our framework uses human tactile measurements to guide both hand--object reconstruction and robot policy learning.
The reconstructed motion is retargeted to the robot and provides the reference for a residual policy trained with motion and contact supervision.

\subsection{Reconstruction from Human Demonstration}
\label{sec:reconstruction}

We record demonstrations using a pair of tactile gloves and an RGB-D camera, which captures both hands and the object from a static first-person viewpoint.
The camera provides color images \(I_t\) and registered depth maps \(D_t\), with calibrated intrinsics \(\mathbf{K}\in\mathbb{R}^{3\times3}\).
For each hand \(h\in\{L,R\}\), the tactile glove records spatially distributed readings \(\boldsymbol{\tau}_t^h\in\mathbb{R}^{N}\), where \(N\) is the number of sensing sites per glove.
We additionally take a metric-scale object mesh \(\mathcal{M}_o\) as input and identify the tabletop, which provides the support surface and the gravity direction (Sec.~\ref{sec:supp_scene}).

\qheading{Visual Motion Initialization}
We use SAM~3~\cite{carion2025sam3}, initialized with instance prompts, to obtain left-hand, right-hand, and object masks, denoted by \(M_t^L\), \(M_t^R\), and \(M_t^o\), respectively.
Given \(\mathcal{M}_o\), \(\{M_t^o,I_t,D_t\}\), and \(\mathbf{K}\), FoundationPose~\cite{wen2024foundationpose} estimates an initial object trajectory \(\{\bar{\mathbf{T}}_t^o\}\).
HaMeR~\cite{pavlakos2024hamer} initializes each hand using the MANO representation~\cite{romero2017mano}, yielding an initial wrist pose \(\bar{\mathbf{T}}_t^h\in\mathrm{SE}(3)\) in the camera frame, finger articulation \(\bar{\boldsymbol{\theta}}_t^h\), and hand shape \(\bar{\boldsymbol{\beta}}_t^h\).
These visual estimates initialize the reconstruction, while the instance masks, measured depth, and tactile readings \(\boldsymbol{\tau}_t^h\) provide observations for subsequent joint refinement.

\qheading{Metric Hand--Object Alignment}
HaMeR's monocular estimates may contain hand-scale and absolute-depth errors.
We first estimate a MANO shape \(\boldsymbol{\beta}\) shared by both hands using their masks and reliable depth observations in a selected frame.
After temporally smoothing the initial trajectories, we optimize only the wrist depth \(z_t^h\) along each initial camera ray.
We also refine the object's 6-DoF trajectory \(\{\mathbf{T}_t^o\}\) by aligning the rendered depth and silhouette of \(\mathcal{M}_o\) with \(D_t\) and \(M_t^o\).
The hand shape and refined object trajectory remain fixed as a metric reference for subsequent contact-guided hand pose optimization.

\qheading{Tactile-Guided Hand Pose Refinement}
Even after metric alignment, hand--object contact geometry may remain ambiguous.
We therefore use tactile observations to constrain contact geometry and regularize relative hand--object motion during sustained contact.
To limit parameter coupling, we jointly optimize only the wrist poses \(\mathbf{T}_t^h\) and finger articulation parameters \(\boldsymbol{\theta}_t^h\) over the sequence using the objective
\begin{equation}
 \begin{aligned}
 \min_{\{\mathbf{T}_t^h,\boldsymbol{\theta}_t^h\}_{t,h}}\quad
 &\mathcal{L}_{\mathrm{visual}}
  +\mathcal{L}_{\mathrm{reg}}
  +\lambda_c\mathcal{L}_{\mathrm{contact}}\\
 &+\lambda_n\mathcal{L}_{\mathrm{normal}}
  +\lambda_p\mathcal{L}_{\mathrm{pres}}.
 \end{aligned}
 \label{eq:objective}
\end{equation}
The visual term \(\mathcal{L}_{\mathrm{visual}}\) combines weighted silhouette and depth losses computed using differentiable rendering~\cite{laine2020nvdiffrast} to maintain agreement with the RGB-D observations.
The regularization term \(\mathcal{L}_{\mathrm{reg}}\) combines weighted terms to encourage temporal smoothness, promote anatomical plausibility, and penalize interpenetration.
The tactile terms \(\mathcal{L}_{\mathrm{contact}}\) and \(\mathcal{L}_{\mathrm{normal}}\) constrain contact geometry, while \(\mathcal{L}_{\mathrm{pres}}\) regularizes temporal changes in the hand's configuration relative to the object.

To connect tactile observations to hand geometry, each hand's readings \(\boldsymbol{\tau}_t^h\in\mathbb{R}^{N}\) are pooled into 54 anatomical regions to form \(\tilde{\boldsymbol{\tau}}_t^h\in\mathbb{R}^{54}\).
The regions refine the contact partition of CPF~\cite{yang2021cpf}, preserving regional contact information at a manageable computational cost.
Each region \(a\) is associated with a tactile anchor on the MANO surface, defined by fixed barycentric coordinates (Fig.~\ref{fig:ego2real}).
The tactile anchor position \(\mathbf{x}_{t,a}^h\) in the object frame is differentiable with respect to wrist pose and finger articulation.

Using this correspondence, we translate detected contact into surface proximity and normal compatibility losses:
\begin{equation}
 \begin{aligned}
 \mathcal{L}_{\mathrm{contact}}
 &= \left\langle
   [-d_{t,a}^h]_+^2+[d_{t,a}^h-\epsilon]_+^2
   \right\rangle_{\mathcal C},\\
 \mathcal{L}_{\mathrm{normal}}
 &= \left\langle
   [\alpha_{t,a}^h-\alpha_0]_+^2
   \right\rangle_{\mathcal C},
 \end{aligned}
 \label{eq:contact_geometry}
\end{equation}
where \(d_{t,a}^h\) is the anchor--object signed distance (positive outside), and \(\alpha_{t,a}^h\) is the deviation from opposing surface normals.
The operator \(\langle\cdot\rangle_{\mathcal C}\) averages over active contacts \(\mathcal C=\{(t,h,a):\tilde\tau_{t,a}^h>\tau_{\mathrm{act}}\}\).
The contact term penalizes penetration and exterior gaps beyond \(\epsilon\), while the normal term penalizes angular deviations beyond \(\alpha_0\).
The distance tolerance \(\epsilon\) is progressively reduced during optimization to tighten contact alignment.

To suppress spurious hand motion relative to the object during sustained contact, we additionally penalize temporal changes in object-frame anchor positions:
\begin{equation}
 \mathcal{L}_{\mathrm{pres}}
 =\sum_{t=2}^{T}\sum_{h,a}\omega_t^h
  \left\lVert
  \mathbf{x}_{t,a}^h-\mathbf{x}_{t-1,a}^h
  \right\rVert_2^2.
 \label{eq:preservation_main}
\end{equation}
The weight \(\omega_t^h\) increases with the temporally smoothed tactile response strength, placing greater emphasis on contact preservation when tactile evidence is stronger.
To avoid constraining contact transitions, it is active only when the hand exhibits contact in both adjacent frames.

\subsection{Human-to-Robot Motion Retargeting}
\label{sec:robot_learning}
We retarget the refined human motions by preserving hand--object interaction geometry rather than directly matching joint angles. Following OmniRetarget~\cite{omniretarget}, TopoRetarget~\cite{toporetarget}, and REGRIND~\cite{regrind}, we construct an interaction mesh~\cite{ho2010interaction} from corresponding hand keypoints and sampled object surface points, and match its Laplacian coordinates across embodiments. For each hand $h$ and frame $t$, we solve
\begin{equation}
\begin{aligned}
\mathbf{z}_{h,t}^{\star}
&=\underset{{\mathbf{z}\in\mathcal F_{h,t}}}{\arg\min}\left\lVert\boldsymbol\ell_{h,t}^{R}(\mathbf{z})
-\boldsymbol\ell_{h,t}^{H}\right\rVert_{W}^{2}
+\lambda_{\mathrm{sm}}
\left\lVert\mathbf{z}-\mathbf{z}_{h,t-1}^{\star}\right\rVert_2^2,
\end{aligned}
\label{eq:retarget}
\end{equation}
where $\mathbf{z}$ comprises the robot wrist pose and actuated finger joint angles; $\boldsymbol\ell^H$ and $\boldsymbol\ell^R$ are the human and robot Laplacian coordinates. The feasible set $\mathcal F_{h,t}$ enforces joint limits, passive-joint coupling, quaternion normalization, and collision constraints. The weighting matrix $W$ and $\lambda_{\mathrm{sm}}$ balance interaction preservation and temporal smoothness. We keep the object trajectory fixed and solve each frame sequentially to obtain the robot motion references for the subsequent residual RL policy learning.

\subsection{Contact Informed Residual RL Policy Learning}
\label{sec:rl}

We learn a residual policy around the retargeted reference trajectories using PPO with an asymmetric actor--critic in mjlab~\cite{mjlab}. 
The robot hand we use is OmniHand 3, which has 10 actuated joints and 6 passive finger joints. The policy adapts reference motions to robot-specific kinematics and contact dynamics while prioritizing object manipulation over exact imitation of human hand configurations.

\qheading{Residual Actions and Observations}
For each hand $h\in\{L,R\}$, the actor predicts wrist translation and rotation residuals together with offsets for actuated finger joints. These actions modify the wrist and joint targets over the retargeting reference, rather than accumulating over time. Wrist targets are executed by impedance control, while finger targets are executed by position control. The actor observes the object pose, short histories of robot wrist and finger states, the previous action, the reference phase, and the corresponding reference wrist and joint targets; its observations include deployment-relevant noise and delay. The critic receives the corresponding clean states, together with privileged fingertip positions, joint velocities, and reference and simulated tactile features. For each hand, the tactile representation contains all binary contact indicators for the fingertips and palm, and the logarithm of the total contact force. Tactile observations are excluded from the actor, enabling tactile-free policy execution.

\qheading{Tracking and Contact Rewards}
We formulate the policy reward to jointly encourage task execution, contact consistency with human demonstrations, and stable robot behavior. The total reward at each control step $t$ is defined as
\begin{equation}
\begin{aligned}
r_t &= \Delta t\left(
r_{\mathrm{tr}}+
r_{\mathrm{con}}+
r_{\mathrm{reg}}
\right), 
\end{aligned}
\label{eq:reward}
\end{equation}
where $\Delta t$ denotes the control interval, and $r_{\mathrm{tr}}$, $r_{\mathrm{con}}$, and $r_{\mathrm{reg}}$ represent the motion tracking, contact supervision, and regularization terms, respectively. 

\textbf{Motion Tracking}
The tracking reward $r_{\mathrm{tr}}$ encourages the robot to reproduce the demonstrated object motion while maintaining approximate agreement with the retargeted wrist trajectories. Specifically, it measures the object keypoint tracking error, object linear and angular velocity errors, and wrist position and orientation errors. Object-centric terms are assigned higher weights to prioritize task completion, while weaker wrist tracking allows embodiment-specific adjustments. In addition, a residual-magnitude penalty in $r_{\mathrm{reg}}$ softly constrains the wrist and finger joint offsets from their reference targets, discouraging unnecessarily large deviations without enforcing strict trajectory imitation. Together, these objectives balance reference motion preservation with the flexibility required to establish physically feasible contacts and accomplish the manipulation task.

\begin{figure*}[!t]
    \centering
    \includegraphics[width=\textwidth]{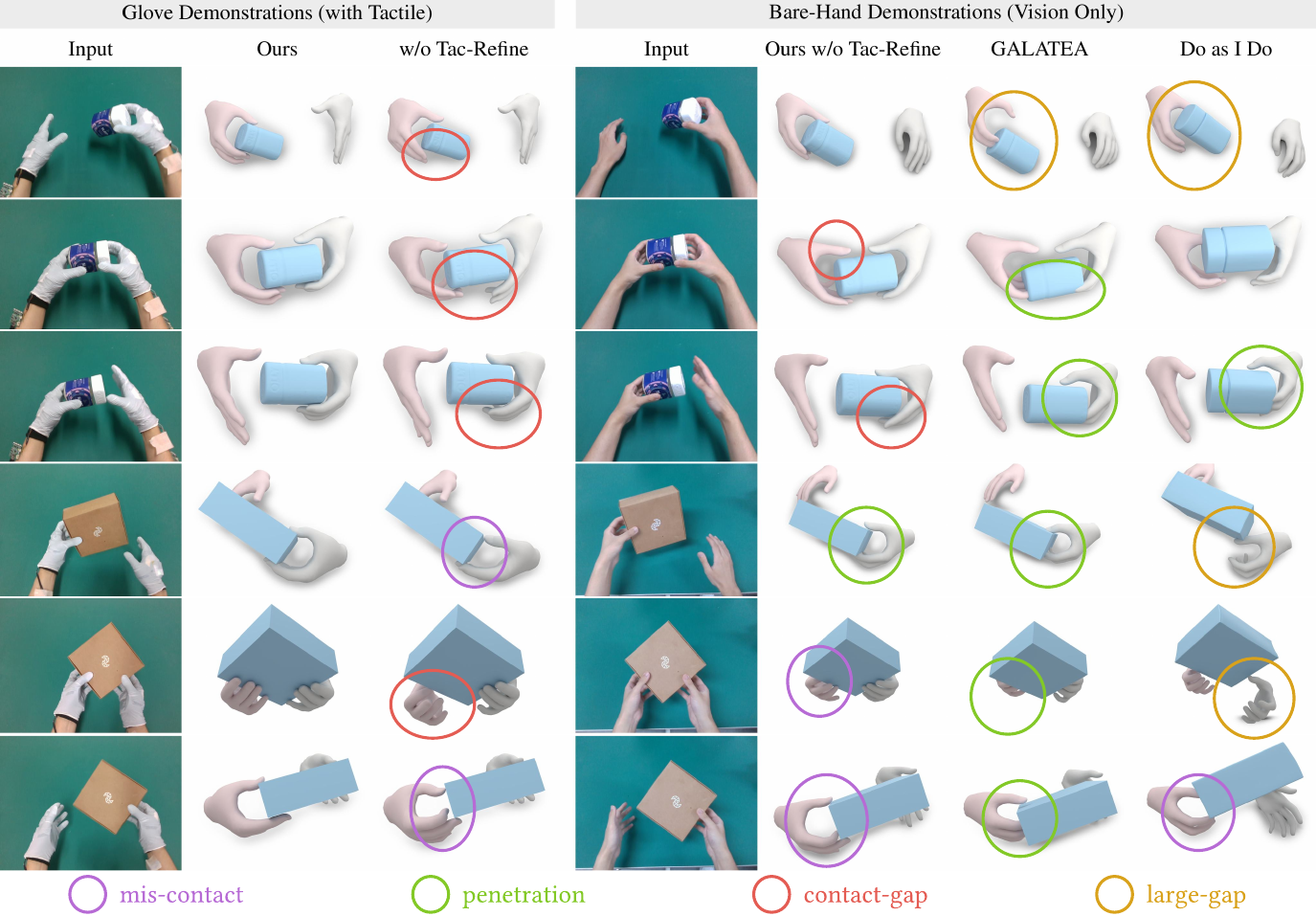}
\caption{\textbf{Qualitative Comparison of HOI Reconstruction.}
Colored circles highlight contact errors.}
    \label{fig:hoirecon-1}
\end{figure*}

\textbf{Contact Supervision}
To transfer human tactile information into robot policy learning, we establish a region-wise contact correspondence between human tactile observations and simulated robot contacts. Specifically, we configure virtual MuJoCo contact sensors over the fingertip pads and palm of the robotic hands, as illustrated in Fig.~\ref{fig:ego2real}. These sensing regions are designed to functionally correspond to the contact-sensitive areas of the human tactile glove, allowing contact information from different embodiments to be represented at a comparable region-wise level. The simulated sensors provide contact-force measurements from which we derive binary contact states and force-related features. Corresponding reference contact states and force summaries are obtained from the synchronized human tactile demonstrations.

Based on this representation, the contact reward $r_{\mathrm{con}}$ consists of the following three complementary objectives:
\begin{equation}
r_{\mathrm{con}} = \sum_{h\in\{L,R\}}
\left(
\lambda_C r_h^C+
\lambda_F r_h^F-
\lambda_P r_h^{\mathrm{pre}}
\right).
\end{equation}
The contact-state reward $r_h^C$ encourages agreement between the simulated and demonstrated contact states of individual contact regions, preserving which hand regions participate in the interaction. The force-similarity reward $r_h^F$ promotes consistency between aggregate simulated and reference contact forces, with tolerance for force discrepancies across embodiments. The premature-contact penalty $r_h^{\mathrm{pre}}$ discourages contact in each region before its first occurrence in the human demonstration, helping the policy respect the demonstrated contact establishment sequence. The coefficients $\lambda_C$, $\lambda_F$, and $\lambda_P$ balance the contributions of these contact-related objectives. Together, these terms provide contact-informed supervision while allowing the robot to adapt its interaction forces and hand configurations to its own physical characteristics.

\textbf{Regularization }
The regularization term $r_{\mathrm{reg}}$ penalizes excessive residual actions, abrupt action changes, violations of action limits, mechanical power consumption, unsafe hand configurations, and failure terminations. These penalties encourage smooth and physically feasible policy execution.

Importantly, simulated contact measurements are used only for training-time rewards and privileged critic observations. The deployed actor does not require tactile observations or robot-side tactile feedback. Detailed reward formulations, weighting coefficients, and contact-sensor configurations are provided in the supplementary material.

\qheading{Termination and Contact-Informed Initialization}
Episodes end at reference completion or timeout, and terminate early when tracking or safety constraints are violated. Only failure terminations incur a penalty. At reset, both hands and the object are initialized from a shared randomly sampled reference frame. To increase exposure to contact establishment, we mix uniform reference-frame sampling with sampling from windows preceding demonstrated contact onsets. Event detection, sampling probabilities, and termination thresholds are detailed in the supplementary material.

\begin{figure}[t]
  \centering
  \begin{subfigure}[b]{0.49\columnwidth}
    \centering
    \includegraphics[width=\linewidth]{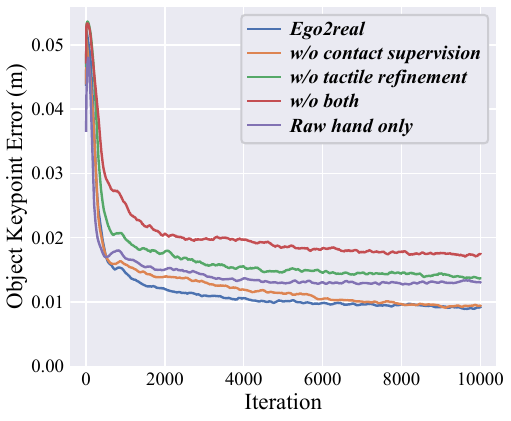}
    \caption{Curve on Bottle Handover task.}
    \label{fig:subim1}
  \end{subfigure}\hfill
  \begin{subfigure}[b]{0.49\columnwidth}
    \centering
    \includegraphics[width=\linewidth]{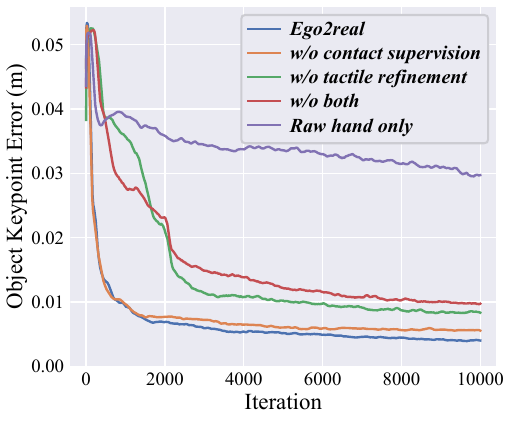}
    \caption{Curve on Box Rotation task.}
    \label{fig:sub2}
  \end{subfigure}

  \caption{\textbf{Object Keypoint Error Curves During Training for Two Tasks.}}
  \label{fig:rl training performance in simulation}
\end{figure}

\begin{figure}[t]
  \centering
    \captionsetup{skip=4pt}
  \includegraphics[width=\linewidth]{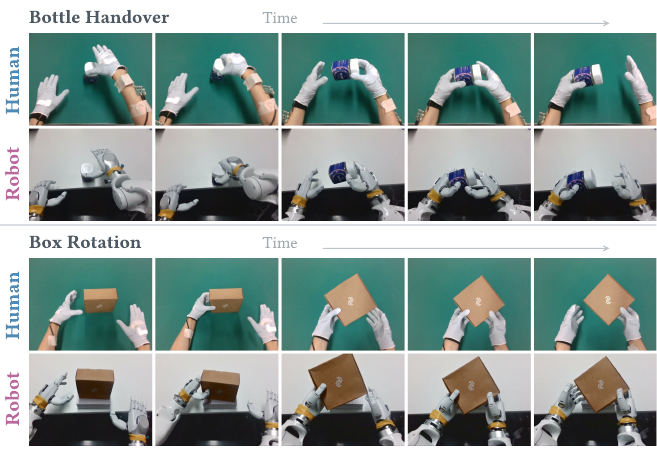}
\caption{\textbf{Real-World Deployment} of Ego2Real on Bottle Handover (top) and Box Rotation (bottom), with human demonstrations and time-aligned robot rollouts.}
  \label{fig:realworld experiment}
\end{figure}

\begin{figure}[t]
  \centering
    \captionsetup{skip=4pt}
  \includegraphics[width=\linewidth]{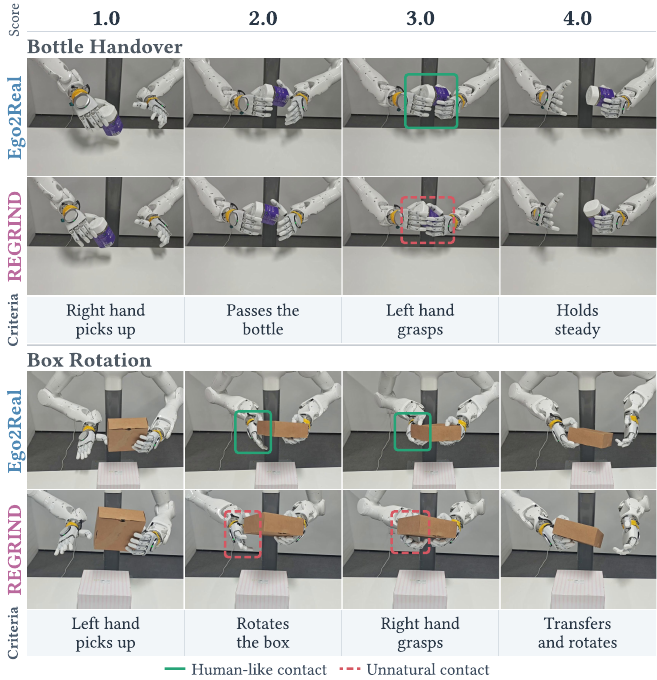}
\caption{\textbf{Qualitative Comparison of Real-World Contact Quality. }Four scoring levels are illustrated for Bottle Handover and Box Rotation. Using identical tactile-refined references, Ego2Real exhibits more human-like contacts than REGRIND, highlighting the benefits of contact-informed RL. Green and red boxes indicate human-like and unnatural contacts, respectively.}
  \label{fig:realcontact quality}
\end{figure}
\section{Experiments}

\subsection{Experimental Setup}

Experiments are conducted on two bimanual manipulation tasks: Bottle Handover and Box Rotation. RGB-D data and tactile measurements are jointly recorded using an Intel RealSense D435i camera at 30 Hz and a pair of tactile gloves with full-palm sensing coverage at 125 Hz, respectively. 
The two streams are synchronized by timestamp, with each visual frame matched to the nearest tactile sample.
Data processing, reconstruction, and policy training use an NVIDIA RTX 4090 GPU. 
The real-world platform consists of Tianji Marvin Pro robotic arms and OmniHand 3 dexterous hands, with FoundationPose~\cite{wen2024foundationpose} providing real-time object pose estimates.

\subsection{HOI Reconstruction}

\label{sec:exp_recon}

We qualitatively evaluate the hand--object interactions reconstructed by Ego2Real, focusing on the physical plausibility of the recovered contacts (Fig.~\ref{fig:hoirecon-1}).
Since GALATEA~\cite{galatea} and Do as I Do~\cite{doasido} are designed for bare-hand videos, we additionally record bare-hand demonstrations of both tasks and compare these methods with our vision-only variant, \emph{Ours w/o Tac-Refine}, which disables the tactile terms ($\lambda_c=\lambda_n=\lambda_p=0$).
Both baselines are extended to the bimanual setting and provided with the same RGB-D input (see the supplementary material).
To isolate the contribution of touch, we compare the full method with \emph{Ours w/o Tac-Refine} on glove demonstrations, using the same inputs and differing only in whether the tactile terms are enabled.

Even with bare hands and measured depth, vision-based reconstruction remains susceptible to contact errors: GALATEA exhibits interpenetration in both tasks and a substantial contact gap in the first Bottle Handover example, Do as I Do produces pronounced hand--object misalignment, and our vision-only variant still leaves contact gaps and misplaced contacts. Visual observations and geometric regularization alone thus do not fully determine the contact configuration.

On glove demonstrations, tactile measurements reduce these ambiguities by specifying which hand regions are in contact. In Bottle Handover, the vision-only variant leaves gaps between the fingers and the bottle or places the fingers at incorrect locations, whereas the full method recovers a grasp with the fingers conforming more closely to the bottle's surface; in Box Rotation, tactile refinement similarly reduces contact gaps and improves the placement of contacting fingers. Because the signals are directly measured, they constrain contact even when the corresponding hand regions are visually occluded.

\begin{table}[t]
    \centering
    \small
    \setlength{\tabcolsep}{3pt}
    \begin{tabular}{@{}lcccc@{}}
        \toprule
        \multirow{2}{*}{\textbf{Method}}
        & \multicolumn{2}{c}{\textbf{Bottle Handover}}
        & \multicolumn{2}{c}{\textbf{Box Rotation}} \\
        \cmidrule(lr){2-3} \cmidrule(lr){4-5}
        & Err.\,(mm) $\downarrow$ & SR\,(\%) $\uparrow$
        & Err.\,(mm) $\downarrow$ & SR\,(\%) $\uparrow$ \\
        \midrule
        Ego2Real   & $4.73_{\pm 0.01}$ & $\mathbf{100.0}_{\pm 0.0}$ & $2.29_{\pm 0.01}$ & $\mathbf{100.0}_{\pm 0.0}$ \\
        REGRIND    & $\mathbf{3.98}_{\pm 0.00}$ & $\mathbf{100.0}_{\pm 0.0}$ & $\mathbf{1.99}_{\pm 0.01}$ & $\mathbf{100.0}_{\pm 0.0}$ \\
        SPIDER     & $12.93_{\pm 0.09}$ & $0.0_{\pm 0.0}$ & $11.76_{\pm 0.06}$ & $0.0_{\pm 0.0}$ \\
        MANIPTRANS & $5.70_{\pm 0.00}$ & $68.5_{\pm 0.4}$ & $8.04_{\pm 2.12}$ & $49.6_{\pm 29.1}$ \\
        \bottomrule
    \end{tabular}
    \caption{\textbf{Simulation Performance Comparison on Bimanual Manipulation Tasks.}}
    \label{tab:simulation performance comparison}
\end{table}

\begin{table}[t]
    \centering
    \small
    \renewcommand{\arraystretch}{0.9}
    \setlength{\tabcolsep}{4pt}
    \begin{tabular}{@{}lccccc@{}}
        \toprule
        \multirow{2}{*}{\textbf{Method}}
        & \multirow{2}{*}{\textbf{Tac-Refine }}
        & \multicolumn{2}{c}{\textbf{Bottle Handover}}
        & \multicolumn{2}{c}{\textbf{Box Rotation}} \\
        \cmidrule(lr){3-4} \cmidrule(lr){5-6}
        & & SR $\uparrow$ & Score $\uparrow$ & SR $\uparrow$ & Score $\uparrow$ \\
        \midrule
        Ego2Real      & \checkmark & $\mathbf{12/15}$ & \textbf{3.20} & $\mathbf{15/15}$ & \textbf{4.00} \\
        REGRIND       & \checkmark & $2/15$           & 2.53          & $\mathbf{15/15}$ & \textbf{4.00} \\
        \midrule
        Ego2Real      & $\times$   & $2/15$           & 2.73          & $0/15$           & 2.00 \\
        REGRIND       & $\times$   & $0/15$           & 2.00          & $0/15$           & 2.60 \\
        \midrule
        Raw hand only & --         & $1/15$           & 2.13          & $4/15$           & 1.13 \\
        \bottomrule
    \end{tabular}
    \caption{\textbf{Real-World Performance Comparison on Bimanual Manipulation Tasks.} \emph{Raw hand only} uses bare-hand demonstrations (no tactile glove) and reconstructs the references without the tactile refinement losses.}
    \label{tab:real world experiment comparison}
\end{table}

\subsection{Simulation Results}

\qheading{Effect of Tactile-Refined Demonstrations}
We first investigate the impact of tactile-guided HOI refinement on downstream reinforcement learning. As shown in Fig.~\ref{fig:rl training performance in simulation}, tactile-refined demonstrations facilitate policy learning and reduce the average object keypoint tracking error, suggesting that incorporating measured contact information into HOI reconstruction provides more effective references for downstream robot skill learning.

\qheading{Comparison with Existing Methods}
We compare Ego2Real against three representative human-to-robot dexterous manipulation methods: REGRIND~\cite{regrind}, SPIDER~\cite{spider}, and MANIPTRANS~\cite{maniptrans}. REGRIND combines interaction-preserving retargeting with residual reinforcement learning and serves as the foundation of our RL pipeline, SPIDER employs sampling-based physics-informed optimization to generate dynamically feasible robot trajectories, and MANIPTRANS combines trajectory imitation with residual RL.

As shown in Table~\ref{tab:simulation performance comparison}, Ego2Real achieves a $100\%$ success rate on both tasks, matching REGRIND and outperforming SPIDER and MANIPTRANS in task completion. Ego2Real exhibits slightly higher object tracking errors than REGRIND (4.73 vs. 3.98 mm for Bottle Handover and 2.29 vs. 1.99 mm for Box Rotation), which may reflect the additional contact supervision objective, yet task execution remains reliable with low tracking errors.

\subsection{Real-World Performance}

We evaluate Ego2Real on two real-world bimanual manipulation tasks, Bottle Handover and Box Rotation (Fig.~\ref{fig:realworld experiment}), and compare it with REGRIND under both tactile-refined and unrefined references. As shown in Table~\ref{tab:real world experiment comparison}, tactile-guided HOI refinement substantially improves real-world success: in Box Rotation, both methods achieve a $100\%$ success rate with tactile-refined references, whereas neither succeeds without refinement, highlighting the importance of accurate HOI reconstruction for downstream policy learning.

The benefits of contact-informed RL are particularly evident in the more challenging Bottle Handover task. With the same tactile-refined references, Ego2Real achieves an $80\%$ success rate (12/15), compared with $13.3\%$ (2/15) for REGRIND, and exhibits more human-like grasping and object-enclosure behaviors (Fig.~\ref{fig:realcontact quality}), suggesting that tactile-derived contact supervision helps preserve task-relevant interactions across embodiments. The raw-hand baseline achieves only 1/15 and 4/15 successful trials on Bottle Handover and Box Rotation, respectively. Combining tactile-guided reconstruction with contact-informed policy learning thus enables reliable real-world dexterous manipulation; detailed success criteria are provided in the supplementary material.
\par

\section{Conclusion and Discussion}
\label{sec:conclusion}

We presented Ego2Real, a framework for learning bimanual dexterous manipulation from synchronized egocentric visuo-tactile human demonstrations. Human touch supervises two complementary stages: tactile-guided HOI reconstruction improves the physical consistency of hand--object motion references, while contact-informed reinforcement learning explicitly supervises region-wise contact states, forces, and contact establishment timing.

The benefit of each stage depends on the contact demands of the manipulation task. For tasks where contact-consistent motion references are sufficient for reliable transfer, tactile-guided reconstruction alone can achieve strong performance. In more contact-demanding interactions requiring coordinated grasp enclosure, force application, and contact transitions, additional contact supervision during policy learning provides substantial benefits. These findings suggest that human touch can guide dexterous skill transfer both implicitly through refined motion references and explicitly through contact-aware training objectives.

Ego2Real currently learns policies for individual demonstrations, relies on object meshes and simulation, and has been evaluated on two rigid-object tasks. Future work will explore goal-conditioned policies across demonstrations, more diverse objects and manipulation skills, and robot-side tactile feedback to further improve contact regulation during execution.

\FloatBarrier

\section*{Acknowledgment}
We thank Yongyi Wan and Linfeng Zhao for their support with the embedded systems of OmniHand3, and the team led by Zhenfei Liu for their support with the tactile gloves.

{
    \small
    \bibliographystyle{ieeenat_fullname}
    \bibliography{main}
}

\clearpage
\setcounter{page}{1}
\maketitlesupplementary
\section{Data Collection and Tactile Glove}
\label{sec:supp_data}

We collect human demonstrations using a pair of tactile gloves and an Intel RealSense D435i camera mounted on an external support to capture a first-person view of the workspace.
Both streams are recorded on a laptop, facilitating data collection across different environments.
The camera captures RGB-D images at 30 Hz, while the gloves record tactile signals at 125 Hz.
We resample each glove’s tactile stream to the 30 Hz RGB-D timeline by selecting the nearest tactile sample for each camera frame.
For each hand $h\in\{L,R\}$, the glove contains $N=1123$ taxels distributed across the palm and fingers, yielding raw tactile readings $\boldsymbol{\tau}_t^h\in\mathbb{R}^{N}$ at frame $t$.

We calibrate the glove responses using a six-axis force/torque sensor.
During calibration, the wearer presses the glove against the sensor's contact surface while we record the summed tactile response and the corresponding normal force.
Assuming a uniform force--response relationship across all taxels, we fit a shared affine mapping by least squares on nonzero readings to convert summed tactile responses into estimates of normal contact force, while zero readings are mapped to zero force.

\section{HOI Reconstruction}
\label{sec:supp_hoi}

This section provides additional details of the HOI reconstruction
pipeline in Sec.~\ref{sec:reconstruction}.
We first describe the object mesh and scene frame, then detail
the visual, regularization, and tactile terms used for hand refinement.
During contact-guided hand refinement, the object trajectory and
shared hand shape are held fixed.
We retain the main-paper indices \(t\in\{1,\ldots,T\}\),
\(h\in\{L,R\}\), and \(a\in\{1,\ldots,54\}\) for frames, hands,
and tactile regions, respectively, and write \([u]_+=\max(u,0)\).
Throughout this section, \(\langle\cdot\rangle\) denotes an arithmetic
mean over the indices or set indicated in the subscript.

\subsection{Object Mesh and Scene Frame}
\label{sec:supp_scene}
Given the measured dimensions of each object, an agentic workflow builds the metric-scale object mesh \(\mathcal{M}_o\).
The same workflow identifies the tabletop in the RGB-D observations and reconstructs it as a support plane.
Since the tabletop is horizontal, its normal is aligned with the vertical direction, and we take the opposite of the upward normal as the gravity direction.
We define the world frame with its vertical axis along this normal.
The support plane also serves as the tabletop collision boundary.

\subsection{Visual Consistency}
\label{sec:supp_hoi_visual}

The visual objective combines silhouette and measured-depth agreement:
\begin{equation}
 \mathcal L_{\mathrm{visual}}
 =\lambda_{\mathrm{sil}}\mathcal L_{\mathrm{sil}}
 +\lambda_{\mathrm{depth}}\mathcal L_{\mathrm{depth}}.
 \label{eq:supp_hoi_visual}
\end{equation}
\qheading{Silhouette Agreement}
For each valid frame--hand pair, we combine a soft intersection-over-union
loss between the rendered and observed hand masks with a boundary-distance loss.
The latter averages normalized distances to the observed boundary, weighted
by the rendered boundary response.
We exclude the projected open wrist boundary of MANO and average the sum
of both losses over valid frame--hand pairs to obtain \(\mathcal L_{\mathrm{sil}}\).

\qheading{Depth Agreement}
We apply the Huber loss to the difference between rendered and measured depth.
Valid pixels lie inside the observed hand masks and outside the wrist
exclusion region, with valid measured and rendered depth.
Within each rendered batch, we average these losses over valid pixels
per hand and then over valid hands, weighting batches by frame count to
obtain \(\mathcal L_{\mathrm{depth}}\).
The preceding metric-alignment stage uses stricter depth filtering to
suppress unreliable observations.

\subsection{Geometric and Temporal Regularization}
\label{sec:supp_hoi_regularization}

The regularizer \(\mathcal L_{\mathrm{reg}}\) is the weighted sum of
temporal, anatomical, and collision terms described below.
Their coefficients are listed in Table~\ref{tab:supp_hoi_weights}.

\qheading{Temporal Smoothness}
We penalize squared wrist translational acceleration
\(\mathcal L_{\mathrm{wp}}\), wrist angular acceleration
\(\mathcal L_{\mathrm{wr}}\), and local finger-joint angular acceleration
\(\mathcal L_{\mathrm{fr}}\), computed by finite differences using frame timestamps.
Each term is averaged over the relevant hands, frames, and joints.

\qheading{Anatomical Plausibility}
We use anatomical axes from \texttt{manotorch} and optimize a reduced
24-dimensional hand pose \(\boldsymbol\theta_t^h\).
Bend and spread limits are enforced by projection after each update,
while \(\mathcal L_{\mathrm{anat}}\) regularizes retained thumb twists toward zero.

\qheading{Hand--Object Non-Penetration}
Let \(\phi_o\) be the object signed distance field, positive outside,
and \(\mathcal S_t^h\) a set of hand surface samples expressed in the
object frame.
The non-penetration term is
\begin{equation}
 \mathcal L_{\mathrm{ho}}
 =\left\langle[m_{\mathrm{ho}}-\phi_o(\mathbf y)]_+\right\rangle_{t,h,\,\mathbf y\in\mathcal S_t^h},
 \label{eq:supp_hoi_clearance}
\end{equation}
where \(m_{\mathrm{ho}}>0\) specifies a small safety margin.
An additional augmented-Lagrangian term enforces the per-frame, per-hand
non-penetration constraint \(g_t^h\le0\):
\begin{equation}
 \begin{aligned}
 g_t^h
 &=m_{\mathrm{ho}}-\min_{\mathbf y\in\mathcal S_t^h}\phi_o(\mathbf y),\\
 \mathcal L_{\mathrm{AL}}
 &=\frac{1}{2\rho}\left\langle[\mu_t^h+\rho g_t^h]_+^2\right\rangle_{t,h},\\
 \mu_t^h&\leftarrow[\mu_t^h+\rho g_t^h]_+,
 \end{aligned}
 \label{eq:supp_hoi_al}
\end{equation}
where \(\rho>0\) is the penalty parameter and the multipliers
\(\mu_t^h\ge0\) are initialized to zero (terms constant with respect to
the hand parameters are omitted).
We activate this term in a later stage of optimization and update the
multipliers periodically.

\qheading{Tabletop Collision}
For hand \(h\) at frame \(t\), let \(\bar\delta_t^h\) and \(\delta_{t,\max}^h\)
denote the mean and maximum tabletop penetration depths over all hand
vertices, with zero penetration above the table. The tabletop collision
loss is \(\mathcal L_{\mathrm{table}}
= \langle \bar\delta_t^h + 1.6\,\delta_{t,\max}^h \rangle_{t,h}\).

\qheading{Hand Self-Collision and Inter-Hand Collision}
Using bone-aligned capsules, \(\mathcal L_{\mathrm{caps}}\) averages
positive overlaps between nonadjacent segments within each hand and all pairs across hands.

\subsection{Additional Definitions for Tactile Constraints}
\label{sec:supp_hoi_tactile}

\qheading{From Tactile Regions to Surface Anchors}
For the taxel set \(\mathcal G_a^h\) assigned to region \(a\), we pool
the raw sensor counts as
\(\tilde\tau_{t,a}^h = \sum_{n\in\mathcal G_a^h}\tau_{t,n}^h\).
Each region is bound to a surface anchor \(\mathbf x_{t,a}^h\) defined by
fixed barycentric coordinates on its assigned MANO triangle in the object
frame, ensuring consistent anatomical correspondence as wrist pose and
finger articulation change.

\qheading{Active Contacts and Surface Normals}
We use the active set \(\mathcal C\) defined in Sec.~\ref{sec:reconstruction}.
For each anchor, the distance and normal deviation are
\begin{equation}
 \begin{aligned}
 d_{t,a}^h&=\phi_o(\mathbf x_{t,a}^h),\\
 \alpha_{t,a}^h
 &=\arccos\!\left(
 -(\hat{\mathbf n}_{t,a}^h)^\top
 \hat{\mathbf n}_o(\mathbf x_{t,a}^h)\right).
 \end{aligned}
 \label{eq:supp_hoi_normals}
\end{equation}
The hand normal \(\hat{\mathbf n}_{t,a}^h\) is the outward unit
normal of the anchor triangle, and \(\hat{\mathbf n}_o\) is the
outward unit normal obtained from the object SDF gradient.
Opposing normals therefore have zero deviation.
Both losses are averaged over \(\mathcal C\), and contribute zero when \(\mathcal C\) is empty.

\qheading{Contact-Preservation Weights}
For each region, we subtract \(\tau_{\mathrm{act}}\) from its tactile
response and average the positive part over a 10-frame window.
We map the largest regional average for hand \(h\) through \(\tanh\)
to obtain a bounded weight \(\omega_t^h\).
The weight is zero at \(t=1\) and is applied thereafter only if the hand
has at least one active region in each of frames \(t-1\) and \(t\).
All anchors on the hand share this weight in Eq.~\eqref{eq:preservation_main},
which penalizes frame-to-frame relative motion between each anchor and the object

\subsection{Objective Weights and Key Parameters}
\label{sec:supp_hoi_parameters}

Table~\ref{tab:supp_hoi_weights} summarizes the objective weights and key parameters.
The coefficient \(\lambda_p\) is gradually increased to its listed
value so that contact geometry can adjust before stronger temporal
preservation is applied.
Distances are evaluated in meters and angular residuals in radians.
The coefficients account for the different units and aggregations
of the losses.

\begin{table}[t]
 \centering
 \caption{Objective weights and key parameters for contact-guided
 hand refinement. }
 \label{tab:supp_hoi_weights}
 \begin{tabular}{@{}llr@{}}
  \toprule
  Term / parameter & Symbol & Value\\
  \midrule
  Silhouette & \(\lambda_{\mathrm{sil}}\) & \(10\)\\
  Depth & \(\lambda_{\mathrm{depth}}\) & \(1\)\\
  Wrist translation & \(\lambda_{\mathrm{wp}}\) & \(0.1\)\\
  Wrist / finger rotation & \(\lambda_{\mathrm{wr}},\lambda_{\mathrm{fr}}\) & \(0.05\) each\\
  Anatomy & \(\lambda_{\mathrm{anat}}\) & \(50\)\\
  Hand--object non-penetration & \(\lambda_{\mathrm{ho}}\) & \(50\)\\
  Tabletop collision & \(\lambda_{\mathrm{table}}\) & \(50\)\\
  Capsule overlap & \(\lambda_{\mathrm{caps}}\) & \(50\)\\
  Additional non-penetration & \(\lambda_{\mathrm{AL}}\) & \(0.5\)\\
  Tactile contact & \(\lambda_c\) & \(10^6\)\\
  Tactile normal & \(\lambda_n\) & \(1\)\\
  Contact preservation & \(\lambda_p\) & \(10^4\)\\
  \midrule
  Activation threshold & \(\tau_{\mathrm{act}}\) & \(80\) counts\\
  Contact tolerance & \(\epsilon\) & \(3\to0.2\)\,mm\\
  Normal tolerance & \(\alpha_0\) & \(30^\circ\)\\
  Safety margin & \(m_{\mathrm{ho}}\) & \(0.3\) mm\\
  \bottomrule
 \end{tabular}
\end{table}

\section{Retargeting Details}
\label{sec:supp_retarget}

\begin{figure*}[t]
  \centering
  \includegraphics[width=\textwidth]{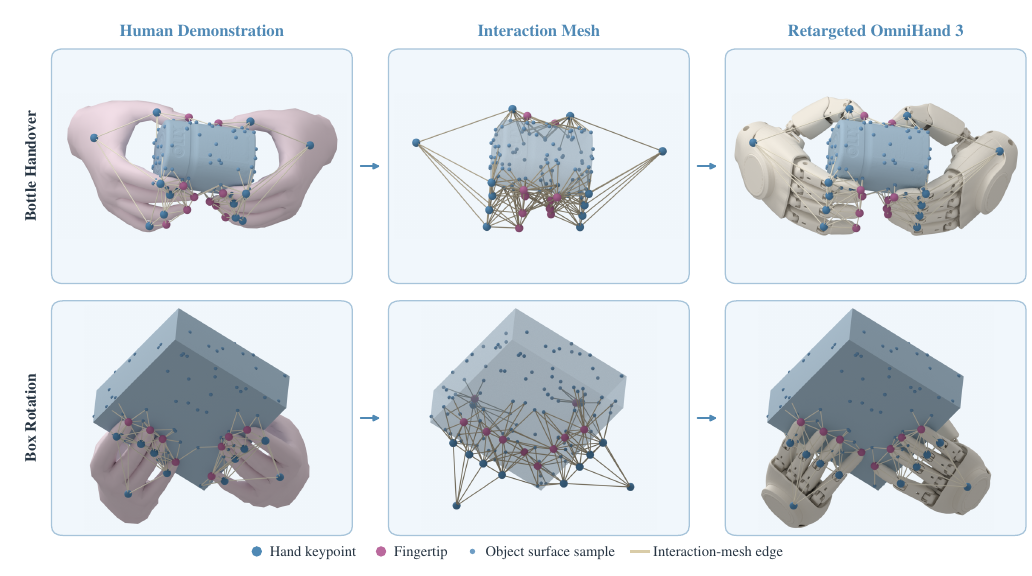}
  \caption{\textbf{Interaction-preserving retargeting for Bottle Handover
  (top) and Box Rotation (bottom).} From left to right, each row shows
  the tactile-refined demonstration with the selected hand keypoints
  (wrist/palm and distal joints in blue, fingertips in magenta) and
  object surface samples, the resulting interaction mesh, and the
  retargeted OmniHand~3 poses. We solve each hand separately along the
  shared object trajectory and overlay both hands for visualization.}
  \label{fig:supp_retarget}
\end{figure*}

Tactile-guided HOI reconstruction captures hand--object contact geometry
in the demonstrated trajectory. Object-agnostic retargeting methods that
match only vectors between hand landmarks, such as DexPilot~\cite{dexpilot},
leave the relative hand--object spatial configuration unconstrained. To
preserve this contact geometry across embodiments, we therefore use a
hand--object interaction mesh that explicitly incorporates the object
surface~\cite{omniretarget,toporetarget,regrind}
(Fig.~\ref{fig:supp_retarget}). For each hand at each frame, we express
the hand keypoints and object surface samples in the object frame and use
Delaunay tetrahedralization to define the mesh edges. Corresponding human
and robot hand keypoints retain the same connections to object surface
samples. Matching the Laplacian coordinates of this interaction mesh thus
preserves the demonstrated hand--object geometric topology and contact
relationships across embodiments.

Constructing this interaction mesh requires selecting hand keypoints with
a consistent kinematic correspondence between the human hand and the
underactuated OmniHand~3. Unlike the three-phalanx structure of human
fingers, each non-thumb finger on OmniHand~3 comprises only two rigid
links with passive joint coupling, creating an inherent morphological
mismatch at the proximal and intermediate joints. We therefore restrict
the correspondence to the wrist and distal finger segments. Specifically,
we map the human wrist to the robot palm to anchor the global hand pose,
and on each finger we match the human distal interphalangeal (DIP) joint
(or interphalangeal joint for the thumb) and fingertip to the robot's
distal joint and fingertip. This distal pair jointly constrains both the
fingertip position and the orientation of the distal link involved in
contact.

On the object side, because the contact geometry shifts across the surface
as the hands transition between grasps, we sample 100 points once per
object, uniformly by area over its entire collision mesh, and keep them
fixed in the object frame. This ensures uniform coverage across face
interiors regardless of mesh tessellation density. With the interaction
mesh constructed from these hand keypoints and object surface samples, we
solve Eq.~\eqref{eq:retarget} via sequential convex programming using
MOSEK.

\section{Reinforcement Learning Details}
This section complements Sec.~\ref{sec:rl} by detailing the policy formulation, the rewards, and the training and evaluation protocols.
Unless noted otherwise, all settings are shared by both tasks and both hands.
Numerical values are collected in
Tables~\ref{tab:supp_rl_reward}--\ref{tab:supp_rl_dr}.

\subsection{Policy Formulation}
\label{sec:supp_rl_policy}

\qheading{Simulation and References}
Both hands and the object are simulated in mjlab with a physics
rate of 120\,Hz and a control rate of 30\,Hz, \ie, a control interval
\(\Delta t=1/30\)\,s.
Each hand is a gravity-compensated floating base whose wrist is driven by an
impedance wrench.
The retargeted motion and the \textcolor{black}{region-wise tactile references} are resampled
onto the control grid, so that reference frame \(k\) provides synchronized
motion and contact targets.

\qheading{Residual Actions}
For each hand \(h\in\{L,R\}\), the actor outputs a wrist translation residual
\(\mathbf a_h^p\), a wrist rotation residual \(\mathbf a_h^R\), and actuated
joint residuals \(\mathbf a_h^q\).
Given the reference wrist pose \((\mathbf p_h^\star,R_h^\star)\) and joint
targets \(\mathbf q_h^\star\) of the current frame, and the clipped action
\(\bar{\mathbf a}=\operatorname{clip}(\mathbf a,-1,1)\), the commanded targets
are
\begin{equation}
 \begin{aligned}
 \mathbf p_h^{\mathrm{cmd}}&=\mathbf p_h^\star+\alpha_p\Delta t\,\bar{\mathbf a}_h^p,\\
 R_h^{\mathrm{cmd}}&=\operatorname{Exp}\bigl(\alpha_R\Delta t\,\bar{\mathbf a}_h^R\bigr)R_h^\star,\\
 \mathbf q_h^{\mathrm{cmd}}&=\operatorname{clip}_{[\mathbf q_{\min},\mathbf q_{\max}]}
 \bigl(\mathbf q_h^\star+\alpha_q\Delta t\,\bar{\mathbf a}_h^q\bigr),
 \end{aligned}
 \label{eq:supp_rl_action}
\end{equation}
where the rates \(\alpha_p\), \(\alpha_R\), and \(\alpha_q\) bound the
deviation from the reference.
Since residuals are added to the time-varying reference rather than
accumulated, a zero action replays the retargeted motion.

\qheading{Observations}
The actor observes the object pose, the wrist poses and actuated joint
positions of both hands over the last two control steps, the previous action,
the reference phase, and the reference wrist and joint targets of the current
frame, with rotations in the continuous 6D representation.
These measured inputs are corrupted by noise and delay
(Sec.~\ref{sec:supp_rl_dr}).
The critic receives the same quantities without corruption, together with
fingertip positions, joint velocities, and tactile features from both the
demonstration and the simulation.
\textcolor{black}{For each hand and source, the tactile feature contains
the binary contact states of the five fingertips and three palm regions
and the logarithm of the total contact force.}
Tactile features are provided only to the critic.

\qheading{Contact Signals}
Each hand has eight object-contact regions \(\mathcal I=\{1,\ldots,8\}\):
five fingertip pads and three palm regions. We average regional summed
normal forces \(F_{h,i}\) and net force magnitudes \(G_{h,i}\) over physics
substeps and set \(m_{h,i}=\mathbb I[G_{h,i}>G_{\mathrm{th}}]\).
As illustrated in Fig.~\ref{fig:ego2real}, we group the tactile anchors
on the MANO surface by anatomical region. Each fingertip is represented
by one aggregate force, and the palm is partitioned into three semantic
regions, each represented by one aggregate force. For hand \(h\) at
reference frame \(k\), let \(\mathcal A_{h,i}\) be the set of MANO anchors
assigned to region \(i\), and let \(f_{h,k,a}\) be the estimated normal
force associated with anchor \(a\). We sum the forces within each region,
\begin{equation}
     F_{h,i}^\star(k)=\sum_{a\in\mathcal A_{h,i}} f_{h,k,a},
 \qquad i\in\mathcal I.
\end{equation}

The proximal and intermediate finger segments differ substantially
between the human hand and the underactuated OmniHand~3, which has fewer
independently articulated finger segments. We therefore omit anchors on
those segments when constructing the RL references and choose regions
that correspond directly to the robot's five fingertip and three palm
sensors. This correspondence lets the reward compare forces and contact
states in regions the simulator can measure directly, while allowing the
robot to use its own finger configurations to establish those contacts.

The reference state \(m_{h,i}^\star\) is determined using a threshold of
\(\eta\) times each region's maximum response in the demonstration.
The tactile feature per hand and source is
\begin{equation}
 \boldsymbol\psi_h=(m_{h,1},\ldots,m_{h,8},\log(1+S_h/F_0)),
 \quad S_h=\sum_{i\in\mathcal I}F_{h,i}.
 \label{eq:supp_rl_tactile_feature}
\end{equation}
For the premature-contact penalty, \(b_{h,i}^\star\)
marks frames before the region's first demonstrated contact and is zero
throughout for regions that never activate. The force \(\widetilde G_{h,i}\)
is the net object-contact force magnitude averaged over physics substeps
for the PIP subtree of a fingertip region, or \(G_{h,i}\) for a palm region.

\subsection{Rewards}
\label{sec:supp_rl_reward}

The reward is \(r_t=\Delta t\,(r_{\mathrm{tr}}+r_{\mathrm{con}}+r_{\mathrm{reg}})\)
as in Sec.~\ref{sec:rl}, with weights and shaping scales listed in
Table~\ref{tab:supp_rl_reward}.

\qheading{Motion Tracking}
Let \(e_o\) be the mean distance between simulated and reference object
keypoints sampled on the object surface, \(e_v\) and \(e_\omega\) the
squared errors of the object linear and angular velocities, and \(e_h^p\)
and \(e_h^R\) the wrist position error and geodesic orientation error:
\begin{equation}
 \begin{aligned}
 r_{\mathrm{tr}}={}&w_o e^{-e_o/\sigma_o}+w_v e^{-e_v/\sigma_v^2}
 +w_\omega e^{-e_\omega/\sigma_\omega^2}\\
 &+w_w\sum_h\bigl(e^{-e_h^p/\sigma_p}+e^{-e_h^R/\sigma_R}\bigr).
 \end{aligned}
 \label{eq:supp_rl_tracking}
\end{equation}
The object terms dominate, whereas the small wrist weight \(w_w\) lets the
policy depart from the retargeted wrist trajectory to establish feasible
contacts.

\qheading{Contact Supervision}
With \(S_h^\star=\sum_{i\in\mathcal I}F_{h,i}^\star\), the three contact terms of
Sec.~\ref{sec:rl} are
\begin{equation}
 \begin{aligned}
 r_h^C&=\frac18\sum_{i\in\mathcal I}\mathbb I\bigl[m_{h,i}=m_{h,i}^\star\bigr],\\
 r_h^F&=\exp\Bigl(-\frac{\bigl[\,|S_h-S_h^\star|-\epsilon_{\mathrm{abs}}-\epsilon_{\mathrm{rel}}S_h^\star\bigr]_+}{\sigma_F\,(S_h^\star+F_0)}\Bigr),\\
 r_h^{\mathrm{pre}}&=\sum_{i\in\mathcal I}b_{h,i}^\star\,\widetilde G_{h,i},
 \end{aligned}
 \label{eq:supp_rl_contact}
\end{equation}
where \([u]_+=\max(u,0)\). The contact supervision reward is
\(r_{\mathrm{con}}=\sum_h(\lambda_C r_h^C+\lambda_F r_h^F-\lambda_P r_h^{\mathrm{pre}})\).
The contact-state term also rewards agreement on non-contact, so regions that
stay off the object in the demonstration are encouraged to do the same.
The force term compares the totals over all eight regions of each hand and ignores
discrepancies within an absolute and a relative tolerance, since contact
forces are not expected to transfer exactly across embodiments; the offset
\(F_0\) keeps it well defined when \(S_h^\star=0\), where it discourages
unnecessary squeezing.
The premature-contact penalty sums force over regions whose demonstrated
contact has yet to begin, so it suppresses forceful
collisions during the approach more than light grazes.

\qheading{Regularization}
We regularize the policy to keep residual actions small and smooth while allowing the adjustments needed to establish contact: 

\begin{equation}
 \begin{aligned}
 r_{\mathrm{reg}}={}&w_a e^{-{\left\langle a_{t,j}^2\right\rangle_j}/\sigma_a^2}
 +w_\Delta e^{-{\left\langle(a_{t,j}-a_{t-1,j})^2\right\rangle_j}/\sigma_\Delta^2}\\
 &+w_b e^{-\sum_j[\,|a_{t,j}|-1]_+/\sigma_b}
 -w_P\sum_hP_h\\
 &
 -w_{LR}\,[\,d_0-d_{LR}]_+
 -w_{\mathrm{fail}}\,\mathbb I_{\mathrm{fail}}.
 \end{aligned}
 \label{eq:supp_rl_reg}
\end{equation}
As in Sec.~\ref{sec:supp_hoi}, \(\langle\cdot\rangle_j\) denotes the arithmetic mean over the indicated index, here it denotes the mean over the action coordinate \(j\). The first three terms favor small residuals, smooth actions, and actions
within their bounds.
The remaining terms penalize the mechanical power \(P_h\) of the finger
actuators and the wrist wrench, wrists closer than \(d_0\), and failure
terminations.

\begin{table}[t]
 \centering
 \small
 \caption{Reward weights and shaping scales. Wrist, contact, and power terms
 are applied to each hand.}
 \label{tab:supp_rl_reward}
 \begin{tabular}{@{}lll@{}}
  \toprule
  Term & Weight & Scale\\
  \midrule
  \multicolumn{3}{@{}l}{\emph{Motion tracking} \(r_{\mathrm{tr}}\)}\\
  Object keypoints & \(w_o=1.5\) & \(\sigma_o=0.015\)\,m\\
  Object linear velocity & \(w_v=1.0\) & \(\sigma_v=1.0\)\,m/s\\
  Object angular velocity & \(w_\omega=1.0\) & \(\sigma_\omega=3.14\)\,rad/s\\
  Wrist position & \(w_w=0.05\) & \(\sigma_p=0.02\)\,m\\
  Wrist orientation & \(w_w=0.05\) & \(\sigma_R=0.2\)\,rad\\
  \midrule
  \multicolumn{3}{@{}l}{\emph{Contact supervision} \(r_{\mathrm{con}}\)}\\
  Contact-state match & \(\lambda_C=0.5\) & --\\
  Force similarity & \(\lambda_F=0.1\) & \(\sigma_F=0.5\)\\
  Premature contact & \(\lambda_P=0.5\)\,N\(^{-1}\) & --\\
  \midrule
  \multicolumn{3}{@{}l}{\emph{Regularization} \(r_{\mathrm{reg}}\)}\\
  Residual magnitude & \(w_a=0.05\) & \(\sigma_a=1.0\)\\
  Action rate & \(w_\Delta=1.0\) & \(\sigma_\Delta=0.5\)\\
  Action bound & \(w_b=1.0\) & \(\sigma_b=1.0\)\\
  Mechanical power & \(w_P=0.001\)\,W\(^{-1}\) & --\\
  Wrist proximity & \(w_{LR}=0.5\)\,m\(^{-1}\) & \(d_0=0.08\)\,m\\
  Failure termination & \(w_{\mathrm{fail}}=10\) & --\\
  \bottomrule
 \end{tabular}
\end{table}

\subsection{Episode Initialization and Termination}
\label{sec:supp_rl_reset}

\qheading{Termination}
An episode ends at the end of the reference or at the time limit.
It terminates early as a failure when the keypoint error exceeds
\(e_{\max}\), the object leaves the workspace, a wrist is farther than
\(d_{\mathrm{far}}\) from the object, the wrists are closer than
\(d_{\min}\), or a fingertip hits the table with a force above
\(F_{\mathrm{gnd}}\).
Only failures incur the penalty \(w_{\mathrm{fail}}\).

\qheading{Contact Informed Initialization}
At reset, both hands and the object are set to the reference state of a
shared frame \(k\), with the passive joints projected onto their couplings.
To expose the policy to contact establishment, we extract contact events from
the demonstration: \textcolor{black}{a hand is in contact when any of its eight reference regions is active}, and an
event is a contact onset, including recontact, that lasts at least
\(T_{\min}\).
Onsets of the two hands closer than \(T_{\mathrm{merge}}\) are merged into one
event.
Each event \(e\) with onset time \(t_e\) defines the pre-contact window
\(\mathcal W_e=[t_e-\tau_{\mathrm{far}},\,t_e-\tau_{\mathrm{near}}]\), and
the reset frame is drawn as
\begin{equation}
 k\sim
 \begin{cases}
  \mathcal U(\mathcal V), & \text{w.p. } 1-p_e,\\
  \mathcal U(\mathcal W_e),\ e\sim\mathcal U(\{1,\ldots,M\}), & \text{w.p. } p_e,
 \end{cases}
 \label{eq:supp_rl_rsi}
\end{equation}
where \(\mathcal V\) contains all non-terminal reference frames and \(M\) is
the number of events.
Sampling events uniformly, rather than frames, keeps short pre-contact phases
from being underrepresented.

\begin{table}[t]
 \centering
 \small
 \setlength{\tabcolsep}{4pt}
 \caption{Control, contact, termination, initialization, and curriculum
 parameters.}
 \label{tab:supp_rl_params}
 \begin{tabular}{@{}llr@{}}
  \toprule
  Parameter & Symbol & Value\\
  \midrule
  Residual rates & \(\alpha_p\) & 1.0\,m/s\\
   & \(\alpha_R,\alpha_q\) & 3.2\,rad/s\\
  Wrist stiffness & \(K_p\) & 1500\,N/m\\
   & \(K_R\) & 30\,N\,m/rad\\
  Wrist damping & \(D_p\) & 100\,N\,s/m\\
   & \(D_R\) & 3\,N\,m\,s/rad\\
  \midrule
  Simulated contact threshold & \(G_{\mathrm{th}}\) & 0.5\,N\\
  Reference activation ratio & \(\eta\) & 5\%\\
  Force tolerances & \(\epsilon_{\mathrm{abs}},\epsilon_{\mathrm{rel}}\) & 0.5\,N, 0.2\\
  Force offset & \(F_0\) & 1\,N\\
  \midrule
  Keypoint error (train, eval) & \(e_{\max},e_{\max}^{\mathrm{eval}}\) & 5\,cm, 3\,cm\\
  Wrist--object distance & \(d_{\mathrm{far}}\) & 0.6\,m\\
  Minimum wrist distance & \(d_{\min}\) & 0.04\,m\\
  Fingertip--table force & \(F_{\mathrm{gnd}}\) & 10\,N\\
  \midrule
  Event-sampling probability & \(p_e\) & 0.5\\
  Min.\ contact duration & \(T_{\min}\) & 0.1\,s\\
  Event merge window & \(T_{\mathrm{merge}}\) & 0.1\,s\\
  Pre-contact window & \([\tau_{\mathrm{near}},\tau_{\mathrm{far}}]\) & \([0.2,0.6]\)\,s\\
  \midrule
  Gravity ramp length & \(n_g\) & \(2{\times}10^4\) steps\\
  Gravity spread & \(\sigma_g\) & 1.5\,m/s\(^2\)\\
  Push activation error & \(e_{\mathrm{push}}\) & 2\,cm\\
  Push ramp length & \(N_{\mathrm{push}}\) & 1667 iterations\\
  Push linear bound & \(v_{\max}\) & \(0.1\to0.5\)\,m/s\\
  Push angular bound & \(\omega_{\max}\) & \(0.2\to1.0\)\,rad/s\\
  \bottomrule
 \end{tabular}
\end{table}

\subsection{Robustness to the Sim-to-Real Gap}
\label{sec:supp_rl_dr}

\qheading{Domain Randomization}
Table~\ref{tab:supp_rl_dr} lists the randomization.
Physical properties of the hands and the object are sampled per environment.
Per episode, we randomize the passive couplings, a command delay, and a joint
offset of each hand, which model the residual mismatch after identification
and motor zero-calibration errors.
The actor observations are corrupted by uniform noise and by delays that are
shared within each sensing source (the object pose, each wrist, and each
hand's joints), reflecting the real-time pose estimation and the
proprioceptive feedback at deployment.

\qheading{Curricula}
Training starts with reduced gravity: at each reset, an environment samples
its gravity from a Gaussian, truncated to the physical range, whose mean
increases linearly from zero to nominal gravity over the first \(n_g\)
control steps.
Once the training keypoint error first falls below \(e_{\mathrm{push}}\),
random velocity perturbations are applied to both wrists and the object at
intervals of 1--5\,s, with bounds that grow linearly over
\(N_{\mathrm{push}}\) iterations.
Both curricula are disabled at evaluation.

\begin{table}[t]
 \centering
 \small
 \caption{Domain randomization. \(\pm x\) denotes \(\mathcal U(-x,x)\), and
 \(\times\) denotes scaling of the nominal value.}
 \label{tab:supp_rl_dr}
 \begin{tabular}{@{}ll@{}}
  \toprule
  Parameter & Range\\
  \midrule
  \multicolumn{2}{@{}l}{\emph{Per environment}}\\
  Hand / object friction & \(\mathcal U(0.5,1.5)\)\\
  Hand mass & \(\times\,\mathcal U(0.95,1.05)\)\\
  Object mass & \(\times\,\mathcal U(0.8,1.2)\)\\
  Object COM offset (\(x,y\) / \(z\)) & \(\pm5\) / \(\pm20\)\,mm\\
  Finger \(k_p\), \(k_d\) & \(\times\,\mathcal U(0.9,1.1)\)\\
  \midrule
  \multicolumn{2}{@{}l}{\emph{Per episode and hand}}\\
  Passive coupling coefficients & \(\times\,\mathcal U(0.7,1.1)\)\\
  Finger command delay & \(\mathcal U(0,75)\)\,ms\\
  Finger joint offset & \(\pm2^\circ\)\\
  \midrule
  \multicolumn{2}{@{}l}{\emph{Actor observations}}\\
  Observation delay & \(\mathcal U(0,67)\)\,ms\\
  Object pos.\ / rot.\ noise & \(\pm1\)\,cm / \(\pm0.04\)\,rad\\
  Wrist pos.\ / rot.\ noise & \(\pm1\)\,cm / \(\pm0.02\)\,rad\\
  Joint position noise & \(\pm0.03\)\,rad\\
  \bottomrule
 \end{tabular}
\end{table}

\subsection{Training and Evaluation}
\label{sec:supp_rl_ppo}

\qheading{Training}
We train with PPO~\cite{schulman2017ppo} for \(10^4\) iterations, using
4096 parallel environments, 24
rollout steps per environment, 5 epochs of 4 mini-batches, an adaptive
learning rate initialized at \(10^{-3}\) with a target KL divergence of 0.01,
a discount of 0.998, a GAE parameter of 0.95, a clip ratio of 0.2, and an
entropy coefficient of 0.002.
The actor and critic are separate MLPs with hidden sizes
\((1024,512,256,128)\) and ELU activations, and inputs are normalized with
running statistics.
The actor's output layer is initialized to zero, so training starts from
replaying the retargeted reference.
Each policy is trained on a single RTX 4090 GPU.

\qheading{Simulation Evaluation}
For Table~\ref{tab:simulation performance comparison}, each policy executes
its mean action from the first reference frame in 1024 parallel
environments, with the physics randomization, observation noise, and
observation delay retained.
A rollout succeeds if it completes the reference without failure, using the
training failure conditions with the stricter keypoint threshold
\(e_{\max}^{\mathrm{eval}}\) (3cm) and without the table-contact condition.
The tracking error is the time-averaged keypoint error \(e_o\) of each
rollout, averaged over environments.
We report the mean and standard deviation over five evaluation seeds of the
same trained policy.

\section{Real-World Experiment Details}
This section supplements the Real-World Performance subsection in the main paper with details on robot hand identification, policy deployment, and the trial scores underlying Table~\ref{tab:real world experiment comparison}.
\subsection{Robot Hand Identification}
\label{sec:supp_sysid}

OmniHand~3 has 10 actuated joints driven by position actuators,
and 6 passive joints (the thumb
PIP and DIP and the four non-thumb DIP joints) that follow their driving joint
through polynomial couplings \(q_{\mathrm p}=\sum_{k=1}^{4}c_k q_{\mathrm a}^k\).
For such underactuated fingers, whether and when a fingertip reaches the
object depends on the actuator response and on the coupling.
We therefore calibrate both against the physical hand before policy learning.

\qheading{Data}
We attach motion-capture rigid bodies to the palm and the finger links and
excite one actuated joint at a time with a chirp command, which covers a
broad frequency range in a single recording.
Each joint angle is recovered from the relative rotation between the adjacent
rigid bodies about their principal rotation axis, and is synchronized with the input position
command.
Sinusoidal commands at several frequencies are held out for validation.

\qheading{Actuator Identification}
Following PACE~\cite{bjelonic2026pace}, we identify, for each actuated joint,
the actuator gains and the command delay \(\boldsymbol\theta=(k_p,k_d,d)\) by
replaying the recorded command in simulation and minimizing the joint-angle
tracking error
\begin{equation}J(\boldsymbol\theta)=\frac1T\sum_{t=1}^{T}
 \bigl(q_t^{\mathrm{sim}}(\boldsymbol\theta)-q_t^{\mathrm{real}}\bigr)^2 .
 \label{eq:supp_sysid}
\end{equation}
Because \(J\) is evaluated through a full simulation rollout and is not
differentiable in the delay, we minimize it with the derivative-free
CMA-ES~\cite{hansen2016cma}.
To decouple the dynamics from calibration errors, the encoder zero offset is
estimated beforehand from the initial static segment of each recording and
then fixed.

\qheading{Passive Couplings}
From the passive-joint angles measured in the same recordings, we refit the
coupling coefficients by least squares regularized toward their nominal
values, which keeps the polynomials well behaved outside the excited range.
The residual mismatch of the identified model is not eliminated but treated as
uncertainty: the coupling coefficients, actuator gains, and command delays are
randomized around the identified values during training
(Sec.~\ref{sec:supp_rl_dr}).
\subsection{Policy Deployment}

\qheading{Closed-Loop Execution}
We train the policy in simulation at 30\,Hz and run inference and update commands on the physical robot at 15\,Hz. The longer command interval gives the Tianji Marvin Pro arms more time to track each target pose, reducing tracking error. Each update advances the reference by one frame, doubling the execution time relative to the 30\,Hz reference. The wrist and finger residuals in Eq.~\eqref{eq:supp_rl_action} use the training value \(\Delta t=1/30\)\,s, preserving their per-command bounds.

\qheading{Trial Scoring Criteria}
Fig.~\ref{fig:realcontact quality} illustrates four successive milestones for each task. Each trial receives the score of the last milestone completed in order, or 0 if it falls short of the pickup milestone. A score of 4 counts as a success. In Table~\ref{tab:real world experiment comparison}, SR reports the number of successful trials out of 15, and Score is the mean over all 15 trials.

\qheading{Bottle Handover}
\textbf{1:} The right hand grasps the bottle and lifts it clear of the support.
\textbf{2:} The right hand moves the bottle within reach of the left hand while retaining its grasp.
\textbf{3:} The left hand closes its grasp on the bottle while the right hand still supports it.
\textbf{4:} The right hand releases the bottle, leaving the left hand to hold it steady on its own.

\qheading{Box Rotation}
\textbf{1:} The left hand grasps the box and lifts it clear of the support.
\textbf{2:} The left hand performs the initial rotation while keeping the box in its grasp.
\textbf{3:} The right hand grasps the rotated box while the left hand still supports it.
\textbf{4:} The left hand releases the box, and the right hand completes the final rotation while retaining control of the box.

\begin{figure}
    \centering
    \includegraphics[width=\linewidth]{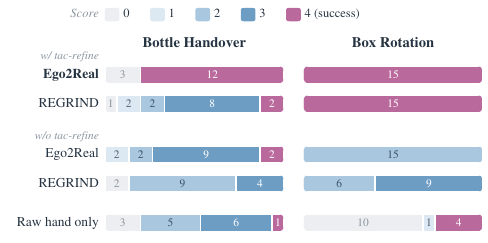}
    \caption{Score Distribution Across Real-World Trials. }
    \label{fig:realworld score details}
\end{figure}

\qheading{Ablation Score Distribution} Fig. ~\ref{fig:realworld score details} shows the distribution of scores across the 15 real-world ablation trials for each experimental setting on Bottle Handover and Box Rotation. For Box Rotation, tactile-guided HOI refinement shifts both Ego2Real and REGRIND from zero successful trials with unrefined references to 15 out of 15 trials at the maximum score. This shared improvement highlights the importance of reference quality. Bottle Handover shows a larger difference between the policies under the same tactile-refined references. REGRIND most frequently scores 3 (8 of 15 trials) and reaches the success score of 4 twice, whereas Ego2Real reaches 4 in 12 of 15 trials. This shift toward the maximum score supports the effectiveness of the contact-informed supervision terms in RL.

\section{Baseline Adaptation for HOI Reconstruction}
\label{sec:supp_baselines}

For the bare-hand comparison in Sec.~\ref{sec:exp_recon}, we run
GALATEA~\cite{galatea} and Do as I Do~\cite{doasido} on the same RGB-D
recordings. For both methods, we replace the depth maps predicted by
MoGe with those measured by the RGB-D camera.
In our bimanual adaptation of GALATEA, each hand has its own rigid pose,
and both hands share a single object pose. We optimize these poses
jointly, summing the per-hand loss terms and counting each object loss
term once. We use the loss weights and optimization schedule from
GALATEA's released configuration for custom videos.
Do as I Do uses HaWoR to reconstruct both hands. We use the right hand
as the scale anchor in both tasks.

\section{Demonstration-to-Robot Rollouts}
\label{sec:supp_rollouts}

\begin{figure*}[t]
  \centering
  \includegraphics[width=\textwidth]{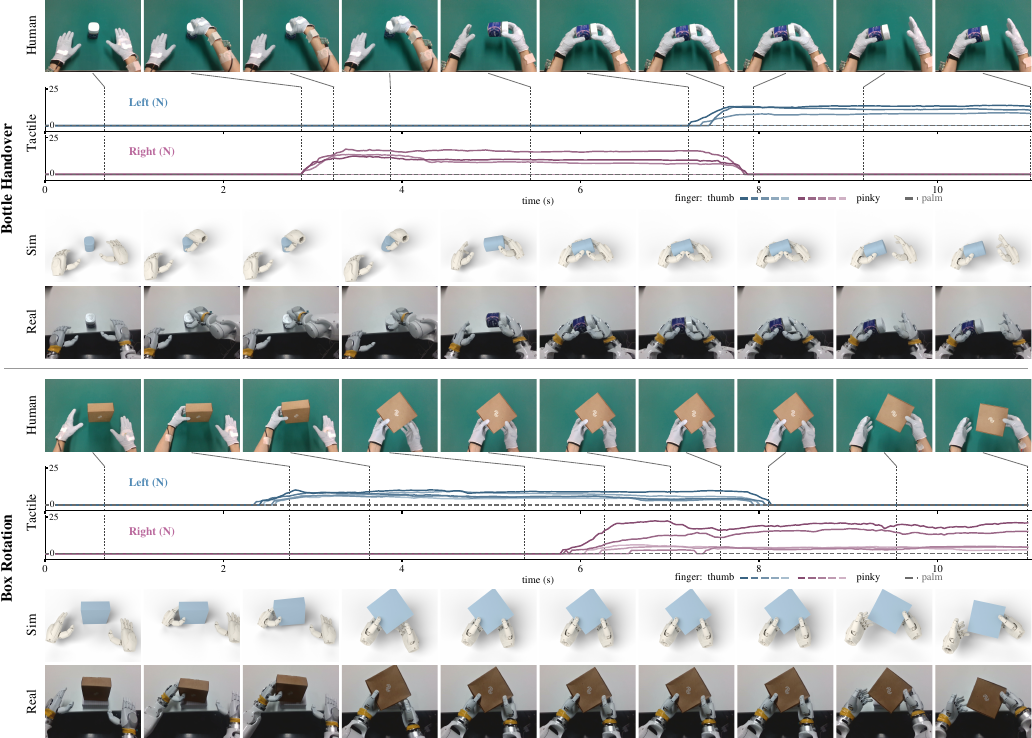}
  \caption{\textbf{From human demonstration to real-world execution} on
  Bottle Handover (top) and Box Rotation (bottom). For each task, the rows
  show the RGB frames of the glove demonstration (\emph{Human}), its
  calibrated normal forces (\emph{Tactile}), the policy rollout in
  simulation (\emph{Sim}), and the real-robot rollout (\emph{Real}). The
  tactile curves give, for the left (blue) and right (pink) hand, the
  calibrated force of each finger from thumb (dark) to pinky (light) as
  solid curves, together with the total palm force (summed over the three
  palm regions) as a dashed curve. Vertical dashed lines mark the ten
  displayed frames, which are sampled more densely around contact changes.
  Sim and Real columns show the same task progress as the demonstration
  frame above them.}
  \label{fig:supp_rollouts}
\end{figure*}

Figure~\ref{fig:supp_rollouts} visualizes, for both tasks, a glove
demonstration together with the corresponding policy rollouts in
simulation and on the real robot. 
The tactile curves show how fingertip forces vary during each human demonstration. In both simulation and real-world execution, the robot exhibits human-like bimanual coordination.

\end{document}